\documentclass[11pt]{article}
\usepackage{geometry}
\usepackage{coling2020}
\usepackage{times}
\usepackage{url}
\usepackage{latexsym}
\usepackage{microtype}
\usepackage{fixltx2e}
\usepackage{url}
\usepackage{graphicx}
\usepackage{siunitx}
\usepackage[bottom]{footmisc}
\usepackage{pdfpages}

\usepackage{float}
\restylefloat{table}

\usepackage{color}
\usepackage{soul}
\usepackage{cleveref}
\crefname{section}{§}{§§}
\Crefname{section}{§}{§§}

\colingfinalcopy 

\title{problemConquero at SemEval-2020 Task 12:  Transformer and Soft label-based approaches}

\author{Karishma Laud$^{*}$\qquad   
	Jagriti Singh\thanks{\quad Authors equally contributed  to this work.}\qquad 
	Randeep Kumar Sahu \qquad  
	Ashutosh Modi  \\
	{Indian Institute of Technology Kanpur (IITK)} \\
	{\tt \{kslaud,jagriti,randeeps\}@iitk.ac.in}  \\
	{\tt ashutoshm@cse.iitk.ac.in}  \\
}
\date{}

\begin{document}
	\maketitle
	\begin{abstract}
		In this paper,
		we present various systems submitted by our team \textit{problemConquero} for SemEval-2020 Shared Task 12 ``Multilingual Offensive Language Identification in Social Media". We participated in all the three sub-tasks of OffensEval-2020, and our final submissions during the evaluation phase included transformer-based approaches and a soft label-based approach. BERT based fine-tuned models were submitted for each language of sub-task A (offensive tweet identification). RoBERTa based fine-tuned model for sub-task B (automatic categorization of offense types) was submitted. We submitted two models for sub-task C (offense target identification), one using soft labels and the other using BERT based fine-tuned model. Our ranks for sub-task A were Greek-19 out of 37, Turkish-22 out of 46, Danish-26 out of 39, Arabic-39 out of 53, and  English-20 out of 85. We achieved a rank of 28 out of 43  for sub-task B. Our best rank for sub-task C was 20 out of 39  using BERT based fine-tuned model. \\
	\end{abstract}

	\section{Introduction}
	
	\blfootnote{
		%
		%
		%
		%
		%
		%
		\hspace{-0.65cm}  
		This work is licensed under a Creative Commons 
		Attribution 4.0 International License.
		License details:
		\url{http://creativecommons.org/licenses/by/4.0/}.
	}
	
	There has been a rise in the use of offensive language on social media platforms. This discourages effective communication, which comes as a drawback against the goal of social media platforms. Hence a need to address the problem of offensive language identification arises. 
	There has been an increasing awareness in communities to filter and moderate tweets. As huge number of users are online, it is difficult to manually moderate each and every content; hence the need for automation arises. As people from different regions, ethnicity interact with each other on online platforms, the detection of multilingual offensive language has become increasingly important. During the training phase of the competition, we started with various machine learning models and proceeded with the model in the evaluation phase, which gave us the highest macro F1 score on the development set. In this paper, we first describe the existing work that has been done for offensive language detection, our proposed approaches, and the results that we obtained. The implementation for our system is made available via Github\footnote{\url{https://github.com/karishmaslaud/OffensEval}}.
	\section{Problem Defintion}
	OffensEval-2020~\cite{zampieri-etal-2020-semeval} consisted of 3 sub-tasks.
	
	\begin{enumerate}
		\item Sub-task A was to classify a tweet as offensive (OFF) or not offensive (NOT). Sub-task A was carried out in five languages: English, Danish, Greek, Turkish, and Arabic.
		
		\item Sub-task B was a classification problem that categorized offensive labeled tweet as targeted (insult or threat) or non-targeted (swear words).
		
		\item Sub-task C was a multi-classification problem that classified the targeted offensive tweet as individual  (IND), group (GRP) or others (OTH).  
	\end{enumerate}
	\section{Related Work}
	Offensive language detection has been studied at various levels of granularity in the form of abusive language detection~\cite{waseem-etal-2017-understanding}, hate speech detection~\cite{schmidt2017survey,kshirsagar-etal-2018-predictive} and cyberbullying~\cite{huang-etal-2018-cyberbullying}. API's like Perspective\footnote{\url{https://www.perspectiveapi.com}} have been developed to detect content toxicity using machine learning models. Various deep learning and ensemble methods have been proposed~\cite{journals/corr/abs-1801-04433} to detect abusive content online. Sentiment based approaches~\cite{brassard-gourdeau-khoury-2019-subversive} have also been used to detect toxicity. 
	
	Besides English, there have been various contributions to the detection of offensive content in other languages. For Greek, \newcite{pavlopoulos-etal-2017-deeper} describe an RNN  based attention model trained on the Gazetta dataset for user content moderation. \newcite{sigurbergsson2019offensive} discuss implementation of  Logistic Regression, Learned-BiLSTM, Fast-BiLSTM  and AUX-Fast-BiLSTM models for Danish. For 
	Arabic, an approach based on convolution
	neural network and bidirectional LSTM has been proposed~\cite{mohaouchane2019detecting}.
	\newcite{abozinadah2017statistical} use a statistical approach for detecting abusive user accounts. \newcite{alakrot2018towards} use an SVM classifier with n-gram features.
	For Turkish, \newcite{ozel2017detection} use Na\"ive Bayes Multinomial, kNN classifier and SVM classifier to detect cyberbullying in which the author used both emoticons and words in the text message as features. Data collection techniques and classical methods in Greek for offensive language have been discussed in  \newcite{pitenis2019detecting}.
	\section{Corpus Description}
	\label{sec:length}
	For OffensEval 2020, datasets were provided by the competition organizers. For sub-task A: data was provided in the form of tweets and their corresponding labels (OFF and NOT) for Arabic~\cite{mubarak2020arabic}, Danish~\cite{sigurbergsson2020offensive}, Turkish~\cite{coltekikin2020}, and Greek~\cite{pitenis2020} languages. For the task on the English Language \cite{rosenthal2020}, for all the three sub-tasks A, B, and C, the data was provided in the form of tweets with their corresponding mean and standard deviation values. The mean and standard deviation scores were confidence measures obtained by semi-supervised learning methods. For sub-task A, the mean score was given to how offensive a tweet was, a higher value indicating a more offensive tweet. For sub-task B, the mean score was given to how close the tweet was to being an untargeted insult (UNT) or how far the tweet was to being a targeted insult (TIN). For sub-task C, a mean score was provided for all the labels ``IND", ``GRP", and ``OTH" with a higher value indicating that the tweet belonged to the given category. 
	
	Multilingual datasets provided to us were unbalanced as follows: Danish consisted of approximately 13\% offensive tweets and 87 percent, not offensive tweets. Similarly, Arabic (20\%: ``OFF"), Turkish (19\%: ``OFF"), and Greek (28\%: ``OFF") training datasets showed a similar ratio imbalance between ``OFF" and ``NOT" labeled tweets. We also made use of the OLID \cite{zampierietal2019} dataset provided in OffensEval 2019 \cite{zampieri2019semeval}, which consisted of the English language tweets and their corresponding labels for sub-tasks A, B, and C conducted in OffensEval 2019. We used this as a validation dataset for sub-task A and error analysis in sub-tasks B and C. During the post-evaluation phase of the competition the organizers also released gold labels, which were actual test labels corresponding to the dataset against which we were evaluated for in the competition.
	
	\section{Sub-task A}
	\subsection{Preprocessing}
	We tried various preprocessing techniques but went with those that gave us the best F1 score on the development set (validation set).  Emoji replacement as a preprocessing technique, as suggested by~\newcite{liu-etal-2019-nuli} in OffensEval 2019, was used for Greek, Arabic, Turkish in sub-task A. We replaced emojis with their corresponding meaning using the emoji library\footnote{\url{https://github.com/carpedm20/emoji}} ~\cite{liu-etal-2019-nuli}. Sentences were preprocessed before passing as input to the BERT model. For Greek, stop words and punctuation were removed by preprocessing with Spacy\footnote{\url{https://spacy.io}}. For Arabic and Turkish consecutive duplicate words like ``@USER @USER" were reduced to a single word ``@USER".
	\subsection{Pre-evaluation Phase}
	During the pre-evaluation phase, we implemented different classical models for Greek and Danish. For Greek and Danish language, we implemented SVM, Logistic Regression, XGBoost, and Random Forest models. We also implemented kNN for Danish.
	\subsection{BERT Model}
	BERT~\cite{devlin2018bert} stands for
	Bidirectional Encoder Representations from
	Transformers released by Google AI\footnote{\url{https://github.com/google-research/bert}}. BERT is a Masked Language Model that jointly conditions on both the left and right context in all the layers of the transformer and hence learns deep bidirectional representations. For sub-task A, English language competition mean scores were mapped to hard labels ``NOT" and ``OFF".  The text data having a mean score greater than 0.5 was labeled as  ``OFF".  We used the BERT-Base Uncased model with a linear classifier for classifying a tweet as offensive or not offensive. The model was trained using Adam optimizer \cite{DBLP:journals/corr/KingmaB14} with weight decay and learning rate of 2e-5. We used a maximum sentence length of 64. We trained the model using an 80:20 training-validation split.
	\subsection{Multilingual BERT Model}
	BERT-Base, Multilingual Cased, is a 12 layer model trained on 104 languages.  
	We used BERT-Base, Multilingual Cased models for Greek, Danish, Turkish, and Arabic language. For Greek, after preprocessing, we did BERT tokenization and  obtained 
	the maximum number of tokens  across all the
	training sentences. We used this information for the truncation of sentences during training and testing. We used a maximum sentence length of 64 for Danish, Arabic and Turkish. We implemented a fine-tuned Multilingual Cased BERT-Base  model with linear classifier using Adam optimizer with weight
	decay (L2 regularization) and a learning rate of 2e-5.
	\subsection{Data Augmentation Techniques}
	In the post-evaluation phase, we translated OLID datasets containing 4400 ``OFF" labeled tweets to Greek and Danish respectively using Google cloud translate\footnote{\url{https://cloud.google.com/translate/docs}} to increase the percentage of ``OFF" tweets in Greek and Danish. The translated datasets are made available via our GitHub repository.
	\section{Sub-task B}
	\subsection{Preprocessing}
	We gave data to the BERT model with and without preprocessing, but as we were getting better results without preprocessing of text sentences, we went ahead with that. We converted the mean probability scores to hard labels ``TIN" and ``UNT" depending on whichever score was maximum. If the mean score was greater than or equal to 0.5, the hard label was ``UNT" else ``TIN" was assigned. After using the above method out of  (188975) soft labeled tweets given for sub-task B, we got approximately 20\% of the tweets classified as ``UNT"  and 0.80\% as ``TIN". We then classified the processed data with the fine-tuned RoBERTa~\cite{DBLP:journals/corr/abs-1907-11692} model.
	\subsection{RoBERTa}
	RoBERTa uses a robustly optimized BERT pre-training approach. It is pre-trained with dynamically masking full sentences without the next sentence prediction
	loss and has a larger byte-level Byte-Pair Encoding. We passed the raw tweets without preprocessing as input to the RoBERTa model. We trained RoBERTa embedding based fined tuned model with a linear classifier using Adam optimizer (with weight decay) and a learning rate of 3e-5. We considered the maximum sentence length of 64.
	\section{Sub-task C}
	\subsection{Preprocessing}
	We used two approaches for sub-task C: BERT and Soft label based approach. 
	For the BERT model, we didn't apply any preprocessing to the given text sentences as our tried combination of preprocessing techniques wasn't working out well while calculating the F1 score on the validation dataset. For the soft labels model, we replaced consecutive duplicate words with a single occurrence of the same word and removed stopwords and punctuation before giving data to the model. Data is provided in form of soft labels and the mean scores that add up to 1.  
	\subsection{BERT}
	We gave the data to the BERT model without applying any preprocessing.In this approach, we converted the soft labels to hard labels. The mean score given in the form of probability for the three labels was converted to ``IND", ``GRP", or ``OTH" depending on which label score was maximum. After using the above method out of 188,973 soft label tweets given for sub-task C, we got approximately 80.7\% of the tweets classified as ``IND", 13.2\% as ``GRP", and 0.61\% as ``OTH". The given data was classified using a fine-tuned BERT model. We trained the BERT model on the entire converted training dataset and used OLID as the validation dataset. The competition organizers provided an all ``IND" baseline dataset during the test phase, which contained the ID of the tweet and all the labels as ``IND". We measured our model's performance on both datasets.
	\subsection{Soft label}
	We used an LSTM \cite{HochSchm97} based approach in which the loss function used the soft labels. Categorical cross-entropy loss is given by $\left[ - \sum_{i=1}^{\mathbf{N}}(\mathbf{p}_i) {\log}(\mathbf{q}_i)\right]$, where $\mathbf{p}_i$ is the mean score of the label  and $\mathbf{q}_i$ is the softmax score calculated for each class via the LSTM, at the end of the every training iteration.
	
	\section{Experiments and Results} \label{sec:experiments}
	
	\subsection{Pre-Evaluation Phase Results}
	The following experiments were conducted during the pre-evaluation phase. For multilingual models, we divided the training data in 80:20 training-validation split, and a batch size of 32 was used. BERT based models were implemented using Pytorch \cite{2019arXiv191201703P} and  Huggingface\footnote{\url{https://github.com/huggingface/transformers/}} Transformers~\cite{Wolf2019HuggingFacesTS} and we have fine-tuned the BERT models using \newcite{1}. For BERT and RoBERTa models, the F1 scores calculated for both pre and post evaluation phases is the average of the F1 scores obtained across batches for a given epoch. Similarly, for BERT and RoBERTa models, the accuracy score calculated is the average of the accuracy scores across batches for a given epoch.
	\\
	\noindent{\textbf{Greek:BERT}}
	We selected the final BERT model via early stopping, which happened around 4 epochs.
	We used a weight decay factor of 0.01 for non-bias and non-normalization layers while fine-tuning the BERT model. We used Adam optimizer with weight decay and a learning rate of 2e-5. We also tried using the average sentence length across the corpus, but it did not lead to improvements in the F1 score hence we went with maximum sentence length. Results for sub-task A: Greek language for both BERT and classical methods based on the validation dataset are presented in Table  \ref{table:12}.
	\\
	\noindent{\textbf{Greek and Danish Classical Methods}}
	We used Scikit-learn ~\cite{JMLR:v12:pedregosa11a} toolkit for implementation. XGBoost\cite{Chen:2016:XST:2939672.2939785} was implemented using the xgboost library\footnote{\url{https://github.com/dmlc/xgboost}}. For Greek and Danish, the data was slightly biased because the Greek and the Danish datasets contained 72 percent and 87 percent, not offensive tweets, respectively. To counter this unbalance in data we used class $weight = balanced$\footnote{\url{https://scikit-learn.org/}} while implementing SVM and Logistic Regression methods. For classical methods (SVM, KNN, XGBoost and Logistic Regression), tf-idf\footnote{\url{https://en.wikipedia.org/wiki/tf-idf/}} unigram based features were used. Results for sub-task A: Danish language for both BERT and classical methods on the validation dataset are presented in Table  \ref{table:12}.
	\\
	\noindent{\textbf{Danish, Arabic, Turkish:BERT}}
	For Arabic and Turkish, we used NLTK toolkit\footnote{\url{https://www.nltk.org}} for tweet tokenization. We used an Adam optimizer with a learning rate of 2e-5 and weight decay optimizer for fine-tuning the BERT model. Result for sub-task A: Turkish language based on BERT fine-tuned model is presented in Table \ref{table:3456} and the result for sub-task A: Arabic language based on BERT based fine-tuned model is presented in Table \ref{table:3456}.
	\begin{table}[H]
		\centering
		\small
		\begin{tabular}{ |c|c|c|c| }  
			\hline
			& Method & F1 Score & Accuracy \\ 
			\hline
			& SVM & 0.638 & 0.82 \\ 
			& Logistic Regression & 0.634 & 0.82 \\ 
			Greek & Random Forest & 0.59 & 0.82\\
			& XGBoost & 0.612 & 0.81\\
			& Bert + Linear layer & 0.81 & 0.85\\
			\hline
			& SVM & 0.73 & 0.90 \\ 
			
			& Logistic Regression & 0.73 & 0.90 \\ 
			
			Danish & K-Nearest Neighbors (k=5) & 0.64 & 0.88\\
			& XGBoost & 0.74 & 0.91\\
			& Bert + Linear Layer & 0.78 & 0.85\\
			\hline
		\end{tabular}
		
		\caption{Accuracy and macro F1 results on the sub-task A: Greek and Danish based on validation set.}
		\label{table:12}
	\end{table}

	\begin{table}[H]
		\centering
		\small
		\begin{tabular}{ |c|c|c|c| }  
			\hline
			& Method & F1 score & Accuracy \\ 
			\hline
			
			Turkish & Bert + Linear Layer & 0.75 & 0.86\\
			\hline
			Arabic & Bert + Linear Layer & 0.82 & 0.90\\
			\hline
			English(Task A) & Bert + Linear Layer & 0.80 & 0.83\\
			\hline
			English(Task B) & Roberta + Linear Layer & 0.85 & 0.91\\
			\hline
		\end{tabular}
		
		\caption{Accuracy and macro F1 results on the sub-task A:Turkish, Arabic, English and sub-task B based on validation set.}
		\label{table:3456}
	\end{table}

	
	\noindent{\textbf{Sub-task A:English}}
	We trained the English data on approx 7,500,000 tweets out of the total soft label tweets given and used the OLID \cite{zampierietal2019} dataset to check F1 macro. As we had resource constraints, we trained 1,000,000 data tweets in one go using a batch size of 32 tweets. We continued this procedure till 7,500,000 tweets. Result is shown in Table \ref{table:3456}.
	
	\noindent{\textbf{Sub-task B}}: For the RoBERTa model, we trained the model for 2 epochs using 80:20 train-val split on the training data. We got the result on the validation set, as indicated in Table \ref{table:3456}. Further training the model for 3 epochs reduced the F1 score on the validation set. 
	Therefore, we selected the model that we obtained after 2 epochs as the final model. We also checked the F1 macro on OLID and an all ``TIN" labeled testing baseline dataset, and we got the F1 macro score as mentioned in Table \ref{table:7}.
	
	
	\begin{table}[H]
		\centering
		\small
		\begin{tabular}{ |c|c|c| }  
			\hline
			Dataset & F1 Score & Accuracy \\ 
			\hline
			All TIN Baseline &  0.466 & 0.875\\
			\hline
			OLID &  0.669 & 0.901\\
			\hline
		\end{tabular}
		\caption{Accuracy and macro F1 results on the  sub-task B based on an all ``TIN" baseline and  OLID dataset.}
		\label{table:7}
	\end{table}
	
	\noindent{\textbf{Sub-Task C:BERT}} For sub-task C, using BERT based approach, we trained the model for 2 epochs. Further increasing the number of epochs decreased the F1 score on the OLID dataset; hence we went for 2 epochs. The results are shown in Table \ref{table:8}.
	
	\begin{table}[H]
		\centering
		\small
		\begin{tabular}{ |c|c|c| }  
			\hline
			Dataset & F1 Score & Accuracy \\ 
			\hline
			All IND Baseline &  0.337 & 0.765\\
			\hline
			OLID &  0.302 & 0.523\\
			\hline
		\end{tabular}
		\caption{Accuracy and macro F1 results on the sub-task C:BERT based on an all ``IND" baseline and OLID dataset.} 
		\label{table:8}
	\end{table}
	
	\noindent{\textbf{Sub-Task C: Soft Labels}} For the soft label based approach for sub-task C, the model was trained for 3 epochs using 90:10 train-val split. Model with best validation loss in the 3 epochs was used. 
	We implemented the model using Keras \cite{2018ascl.soft06022C}. We used one hot encoded vectors embedding with a dimension of 128 as input for the given LSTM model built using \newcite{S3}. The LSTM model had a 1D Spatial Dropout~\cite{2014arXiv1411.4280T} of 0.2 before the LSTM and applied a dropout of 0.2 for the inputs of the LSTM and a recurrent dropout \cite{2015arXiv151205287G} of 0.2. Results are shown in Table \ref{table:9}.
	\begin{table}[H]
		\centering
		\small
		\begin{tabular}{ |c|c|c| }  
			\hline
			Dataset & F1 Score & Accuracy \\ 
			\hline
			Development Set &  0.63 & 0.755\\
			\hline
			All IND Baseline &  0.293 & 0.784\\
			\hline
		\end{tabular}
		\caption{Accuracy and macro F1 results on the sub-task C:Soft Labels approach based on development set and on an all ``IND" baseline.}
		\label{table:9}
	\end{table}

	\subsection{Evaluation Phase Results} 
	Accuracy and macro F1 results on the official test set of  OffenseEval 2020 are presented in Table \ref{table:10}.
	\begin{table}[H]
		\centering
		\small
		\begin{tabular}{ |c|c|c| }  
			\hline
			Sub-task & Language & F1 Score \\ 
			\hline
			A & Arabic & 0.802 \\ 
			\hline
			A & Danish & 0.697 \\ 
			\hline
			A  & Greek & 0.812\\
			\hline
			A  & Turkish & 0.755\\
			\hline
			A  & English-Bert & 0.908\\
			\hline
			B  & English-Roberta & 0.556\\
			\hline
			C  & English-BERT & 0.587\\
			\hline
			C  & English-Soft Label & 0.557\\
			\hline
		\end{tabular}
		\caption{Accuracy and macro F1 results on all sub-tasks based on the test dataset as provided by the competition organizers.}
		\label{table:10}
	\end{table}
	
	\subsection{Post-Evaluation Phase Results}
	
	\textbf{Data Augmentation Results} The F1 score on the test gold labels after data augmentation is presented in Table \ref{table:11} for Greek and Danish languages. We augmented Greek and Danish training datasets with OLID ``OFF" labeled tweets translated to Greek and Danish, respectively. 
	
	\begin{table}[H]
		\centering
		\small
		\begin{tabular}{ |c|c|c| }  
			\hline
			Method & BEST F1 Score & Accuracy \\ 
			\hline
			BERT+Linear Greek  & 0.76 & 0.842 \\ 
			\hline
			BERT+Linear Danish  & 0.724 & 0.878 \\ 
			\hline
			
		\end{tabular}
		\caption{Accuracy and macro F1 results on sub-task A on the test gold labels:Greek and Danish language  trained using  augmented dataset.}
		\label{table:11}
	\end{table}
	\section{Error Analysis}
	\subsection{Quantitative Analysis}
\begin{figure}[H] \centering{
			\includegraphics[scale=0.60]{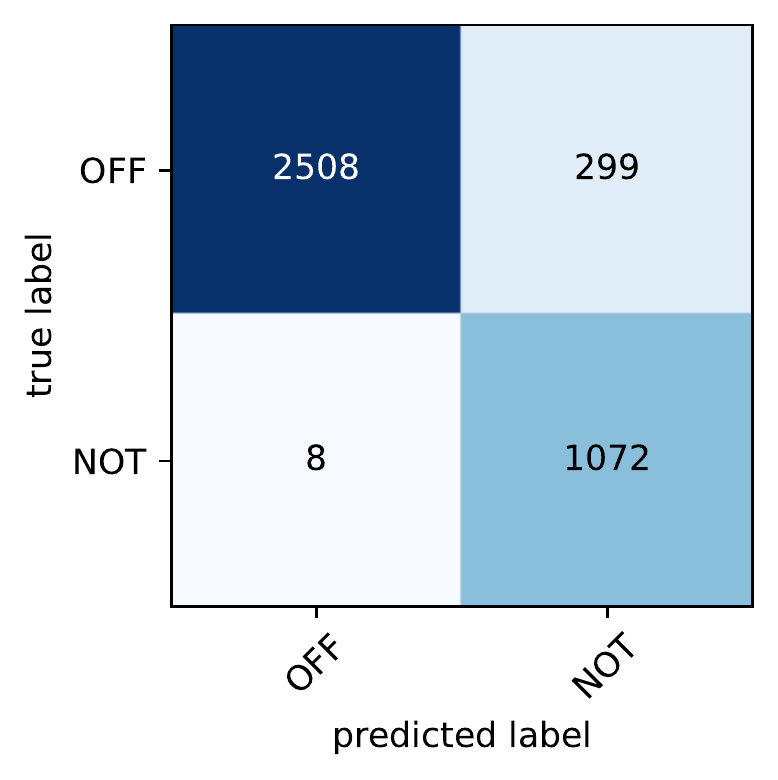}
			\includegraphics[scale=0.60]{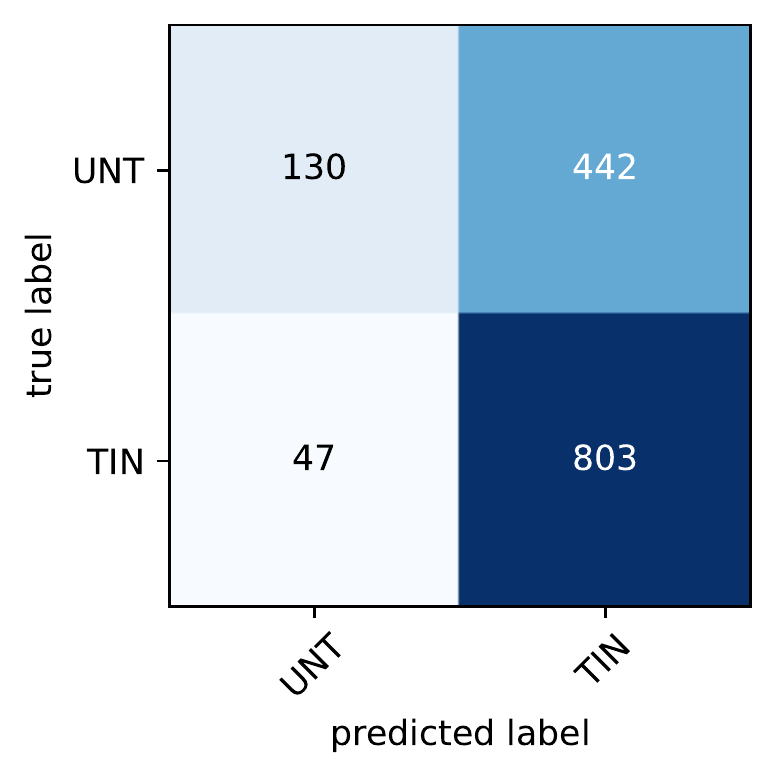}}
		\caption{(a) Confusion Matrix for sub-task A:English  ,\hspace{0.4cm} (b) Confusion Matrix for sub-task B on the\hspace{0.2cm} on the gold   labels dataset  \hspace{5.2cm} gold labels dataset }

		\label{pdf:1}
	\end{figure}

	\noindent\textbf{Data Augmentation} For Greek and Danish ( Section ~\ref{sec:experiments}), the data augmentation techniques using the  OLID data  for Greek and Danish respectively dropped the F1 score on the test dataset for Greek and improved the F1 score for Danish. However, we are not making any claims as we are still working on improving the score on the BERT models, each trained on original and augmented data.\\
	\noindent\textbf{English Language}\\
	The confusion matrices for sub-tasks A, B, and C have been implemented using scikit-learn and the seaborn library \cite{michael_waskom_2017_883859}.\\
	\noindent\textbf{Sub-task A} For the BERT model submitted for English sub-task A,  the confusion matrix for the gold labels is as shown in Figure \ref{pdf:1} (a). We also analyzed the tweets which were offensive but misclassified as not offensive for the gold labels dataset. Some of the tweets didn't have an offensive tone but had incorporated some form of offensive word\footnote{offensive words have been censored while analyzing texts}. E.g.``D**n... it’s true. RIP Toni Morrison". For some of the not offensive tweets that were misclassified as offensive, the tweets had a strong negative tone, but words used in it were not offensive. E.g. ``One thing I hate most is a liar @elongated-nose-emoji".\\ 
	\noindent\textbf{Sub-task B} For the RoBERTa model implemented for sub-task B, the confusion matrix for gold labels is as shown in Figure \ref{pdf:1} (b). We can see that the false-negative rate of ``UNT" (untargeted insult) class was approximately 77\% of the total ``UNT" labels for testing. This served as a major error contributor for the F1 score for sub-task B. For the gold labels, some of the targeted tweets which were misclassified as ``UNT" but were ``TIN" included text sentences that described the target's behavior or situation in an offensive manner. E.g. ``The man eats like a f******g animal." or ``R**e B**y what the f**k". Also, many of the tweets misclassified as ``UNT" by our model included the offensive word ``s**t" used in various connotations. Some of the untargeted tweets that were misclassified as targeted included texts that were expressive of one's feelings or opinions in an aggressive manner or self-criticizing or self-blaming texts. E.g. ``I’m the reason why a lot of s**t is the way it is ." or ``Looking back at old photos of me makes me physically sick like why was I allowed to be f*gly as h**l?? Who let me do that y'all fake as f**k". 
	
	\begin{figure}[H] \centering{
			\includegraphics[scale=0.60]{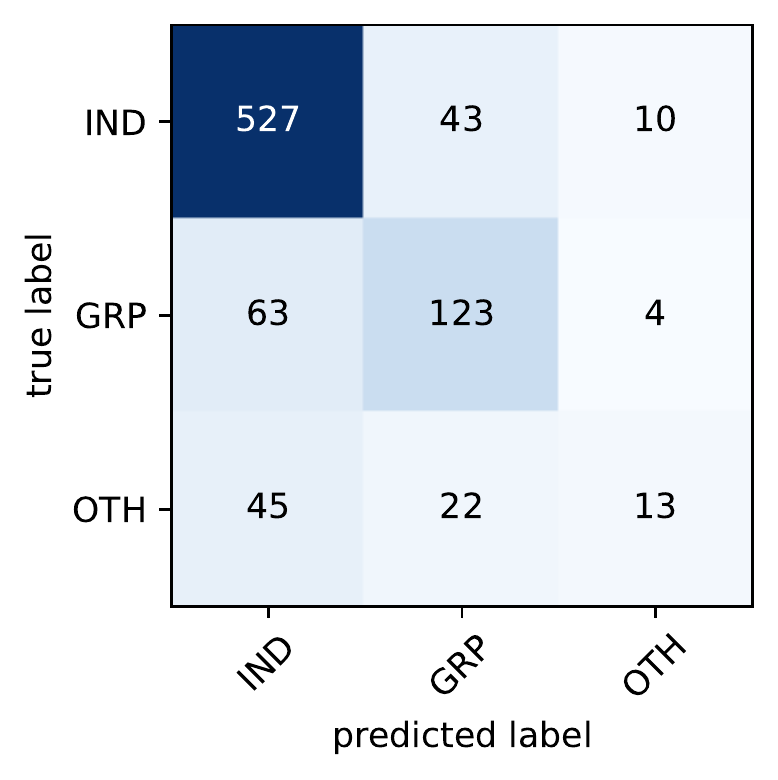}
			\includegraphics[scale=0.60]{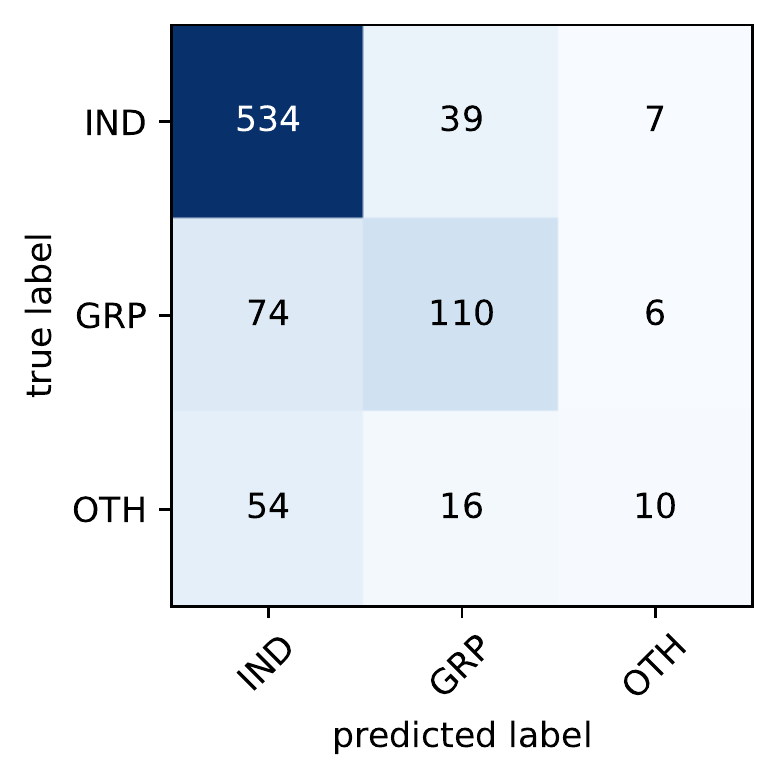}}
		\caption{ (a) Confusion Matrix for sub-task C:Bert  ,\hspace{0.4cm} (b) Confusion Matrix for sub-task C:Soft Labels \hspace{0.2cm} model on the gold   labels dataset   \hspace{3.2cm} Model on the gold labels dataset}
		\label{pdf:2}
	\end{figure}
	
	\noindent\textbf{Sub-task C} The confusion matrix corresponding to the Gold dataset for the BERT model is shown in Figure \ref{pdf:2} (a), and for the soft labels model is shown in Figure \ref{pdf:2} (b). For BERT model, 33\% of the ``GRP" labeled tweets got predicted as ``IND" and for ``OTH"  hardly 16\% of the true labels got classified correctly. Similarly, for the soft labels model, 38\% of the ``GRP" labeled tweets got predicted as ``IND" and for ``OTH"  hardly 12.5\% of the true labels got classified correctly. Here the imbalance in the labeled tweets got propagated to prediction time with ``GRP" and ``OTH" labels getting miss-classified. We also analyzed tweets that were actually ``IND" but misclassified using the BERT model as either ``GRP" or ``OTH" for the gold labeled dataset. Some of the tweets seemed to have a plural entitiy, and an individual was either being insulted in the tweet due to these entities or the individual was being labeled as an offensive plural entity. E.g. ``@USER @USER These children become miserable in your mouth, the truth is they are taken good care of by government.". For the soft labels model, we analyzed the tweets with their true labels as ``IND" but misclassified as ``GRP" or ``OTH". Some of the tweets included the word ``racist" or some forms of racism in the targeted insult for the individual.
	
	

	\section{Future Work}
	\subsection{BERT Bidirectional LSTM}
	During the post-evaluation phase, for Greek, we are working on models that focus on the addition of deep learning models on top of the frozen BERT layer using \newcite{S2}. By addition of  Bidirectional GRU\cite{DBLP:journals/corr/ChoMGBSB14} on top of BERT layer we got an F1 score of 0.833. By the addition of Bidirectional LSTM (Bi-LSTM) on top of the BERT layer, we got an F1 score of 0.797. After concatenation of the last 4 hidden layers of BERT for both  Bidirectional GRU(Bi-GRU) and Bi-LSTM models, the F1 score was 0.819 and 0.837, respectively. The results were obtained after training each model for a maximum of 10 epochs and storing the model with the largest F1 score among the 10 epochs. These results presented in Table \ref{table:14} are based on the test gold labels, and the F1 and accuracy scores are obtained by averaging across batches in an epoch ( Section ~\ref{sec:experiments}). We are still in the process of tuning the hyperparameters for getting a stable output as we are experiencing a lot of deviation in our F1 scores. We have given the best F1 score obtained after the training of the model. We are also planning to take care of the deviation using an ensemble method with different dropout rates.
	\begin{table}[H]
		\centering
		\small
		\begin{tabular}{ |c|c|c| }  
			\hline
			Method & F1 Score & Accuracy \\ 
			\hline
			BERT+Bi-GRU & 0.833 & 0.906 \\ 
			\hline
			BERT+Bi-GRU+ BERT Hidden Layers Concatenated  & 0.819 & 0.915 \\
			
			\hline
			BERT+Bi-LSTM & 0.797 & 0.883 \\ 
			\hline
			BERT+Bi-LSTM + BERT Hidden Layers Concatenated  & 0.837 & 0.915\\
			\hline
		\end{tabular}
		\caption{Accuracy and macro F1 results on the sub-task A: Greek based on different techniques}
		\label{table:14}
	\end{table}
	
	\subsection{Data Augmentation Techniques}
	We are planning on trying other data augmentation techniques in addition to the ones proposed. One of them considers using back translations (English to language X to English, Greek to language Y to Greek) as suggested by~\newcite{aggarwal-etal-2019-ltl} in their paper for English sub-tasks for OffensEval 2019. We are planning to use these methods for sub-tasks A and C to improve the macro F1 score.
	\section{Conclusion}
	In this paper, we described the approaches that we used for sub-tasks A, B, and C. We used Transformer based approaches for all sub-tasks and a soft label LSTM based approach for sub-task C. We also described our future approaches based on BERT embedding with Bi-GRU and Bi-LSTM. We are planning to use the data augmentation techniques and try to improve the F1 score on the gold labels for each sub-task. We made use of google colab \footnote{\url{https://colab.research.google.com/}} while developing our project, and the notebooks are made available at our GitHub page. 
	\bibliographystyle{coling}
	\bibliography{semeval2020}
\end{document}